\documentclass[11pt, oneside]{article}
\usepackage[affil-it]{authblk}
\usepackage{hyperref}

\usepackage{tabularx}
\usepackage{color,soul}
\usepackage{mathtools}
\usepackage[vlined]{algorithm2e}
\SetKwInput{KwInput}{Input}
\SetKwInput{KwOutput}{Output}
\SetKwFunction{GenerateCode}{GenerateCode}
\title{Self-Modifying Code in \\Open-Ended Evolutionary Systems}
\author{Patrik Christen\footnote{Correspondence to: Patrik Christen, \href{mailto:patrik.christen@fhnw.ch}{patrik.christen@fhnw.ch}, FHNW University of Applied Sciences and Arts Northwestern Switzerland, Institute for Information Systems, Riggenbachstrasse 16, 4600 Olten, Switzerland. An earlier and shorter version of this manuscript was presented at The Fourth Workshop on Open-Ended Evolution (OEE4), The 2021 Conference on Artificial Life (ALife 2021) \cite{Christen.2021}.}}
\affil{FHNW, Institute for Information Systems, Switzerland}
\date{28 February 2022}
\begin{document}
\maketitle
\begin{abstract}
Having a model and being able to implement open-ended evolutionary systems is important for advancing our understanding of open-endedness. Complex systems science and newest generation high-level programming languages provide intriguing possibilities to do so. First, some recent advances in modelling and implementing open-ended evolutionary systems are reviewed. Then, the so-called allagmatic method is introduced that describes, models, implements, and allows interpretation of complex systems. After highlighting some current modelling and implementation challenges, model building blocks of open-ended evolutionary systems are identified, a system metamodel of open-ended evolution is formalised in the allagmatic method, an implementation self-modifying code prototype with a high-level programming language is provided, and guidance from the allagmatic method to create code blocks is described. The proposed prototype allows modifying code at runtime in a controlled way within a system metamodel. Since the allagmatic method has been built based on metaphysical concepts borrowed from Gilbert Simondon and Alfred N. Whitehead, the proposed prototype provides a promising starting point to interpret novelty generated at runtime with the help of a metaphysical framework.
\end{abstract}
\section{Introduction}
The diversity and complexity of organisms created by biological evolution over the last billions of years is staggering. It seems to really be a never-ending story which lies at the ground of the invention of all nature and continues to add more and more inventions \cite{Stanley.2019,Stanley.2017}. Engineered physical systems, evolutionary and genetic algorithms, artificial intelligence, deep learning, and other computational methods are thus far from simulating and explaining diversity, creativity, and open-endedness exhibited by biological evolution. The main deficiency is that all these computational systems reach an equilibrium state at some point from which they are not capable of generating anything new -- they are essentially cul-de-sacs.

Understanding the open-endedness of biological evolution is a grand challenge, considered one of the ``millennium prize problems'' \cite{Bedau.2000} in the field of artificial life. If implemented in an open-ended computational system, it would have major implications far beyond artificial life \cite{Stanley.2019,Stanley.2017}. It would allow us to invent virtually everything including new architectures, furniture, cars, games, and of course algorithms and software in general \cite{Stanley.2019,Stanley.2017}. It would most likely bring us closer to strong artificial intelligence, since only biological evolution has created it so far \cite{Stanley.2019,Stanley.2017}.

Furthermore, open-endedness has been observed in various complex systems such as human languages, legal systems, economic and financial systems, and technological innovation showing its relevance as well as urging its study \cite{Bedau.2019,Banzhaf.2016}. These systems are an important part of our society. Better understanding of their open-ended dynamics, i.e., when a system is completely reorganising itself from time to time (i.e., it crashes), is key to managing them. This, in turn, feeds back onto some of our biggest challenges including climate change and socio-economic stability \cite{Thurner.2020}.

This paper first presents a short review of some recent advances in modelling and implementing open-ended evolutionary systems from the fields of artificial life including the open-ended evolution community, complex systems science, artificial intelligence including evolutionary algorithms, and evolutionary biology. Then, the so-called allagmatic method is introduced that describes, models, implements, and allows interpretation of complex systems. After highlighting some current modelling and implementation challenges, model building blocks of open-ended evolutionary systems are identified, a system metamodel of open-ended evolution is proposed as part of the allagmatic method, a self-modifying code prototype in a high-level programming language is presented, and finally the allagmatic method is used as guidance to create code blocks with the developed self-modifying code prototype. It is concluded that the proposed approach provides a promising starting point to interpret novelty generated at runtime.

\section{Recent Advances in Modelling Open-Ended Evolutionary Systems}

\subsection{Definitions}
Although progress has been made, especially by the open-ended evolution community, much remains to be explored \cite{Packard.201955}. Before having a closer look at modelling, we start with some preliminaries regarding the definition of open-ended evolution and open-endedness.

\textit{Open-endedness} has been defined as the ability to continually produce novelty and/or complexity whereby novelty is classified as variation, innovation, and emergence \cite{Banzhaf.2016}. Based on creativity research \cite{Boden.2015}, different terms for this classification were suggested, namely exploratory, expansive, and transformational novelty, respectively \cite{Taylor.2019}. The latter terms will be used here to avoid interpretation issues with innovation and emergence. Regardless of the terminology, both definitions relate to a formal model and metamodel of the system under study. \textit{Exploratory novelty} can be described using the current model, \textit{expansive novelty} requires a change in that model but still uses concepts in the metamodel, and \textit{transformational novelty} introduces new concepts necessitating a change in the metamodel \cite{Banzhaf.2016,Taylor.2019}. With their connection to model and metamodel, they provide a way to determine whether and which kind of novelty emerges in an open-ended evolutionary system.

Defining \textit{complexity} and its measurement in open-ended evolutionary systems is a topic of ongoing research too. Dolson et al. \cite{Dolson.2019} recommend an information-theoretic approach based on the count of informative sites across all components in a population and suggest improving it by also accounting for all possible mutations and by considering epistatic interactions. Channon \cite{Channon.2019} defines individual complexity as the diversity of adaptive components in the individual, i.e., the number of active genes. Furthermore, in evolutionary biology, information is quantified with respect to different sources available to an adapting organism, from ancestors and the environment \cite{Rivoire.2016}.

\subsection{Modelling Contributions from Artificial Life and Open-Ended Evolution Community}
Banzhaf et al. \cite{Banzhaf.2016} have argued that open-endedness in physical systems such as in a computation are hard to prove in a finite universe and therefore one might look for producing a sufficient rather than an infinite number of open-ended events, which is then called \textit{effectively} open-ended. To achieve this despite the limits on computational power, it has been suggested to hard-code certain elements of the model, e.g., the process of replication, into so-called \textit{shortcuts} \cite{Banzhaf.2016,Taylor.2019}. Taylor \cite{Taylor.2019}  bases shortcuts on generally accepted processes of Darwinian evolution: phenotype generation (from the genotype), phenotype evaluation, and reproduction with variation. Ongoing evolutionary activity and with that exploratory open-endedness is promoted by modifying the adaptive landscape, the topology of genetic space, or the genotype-phenotype map. He further argues that none of these expand the phenotype-space itself and thus do not help us for expansive and transformational open-endedness, where so-called \textit{door-opening} states in phenotype-space are needed. The complexity of physical and chemical laws provides a vast space for biological systems whereas in computational systems one might dynamically increase the space instead, e.g., providing access to additional resources on the Internet \cite{Boden.2015,Taylor.2019,Taylor.20164sn}. In contrast, at the third workshop on open-ended evolution, Taylor and others from the open-ended evolution community mentioned that current computational systems implement rather scanty environments and organisms.

Taylor \cite{Taylor.2019} also proposes two possible intrinsic mechanisms to access new states. The first is via \textit{exaptation}, where a trait changes its function to a different one from the one it was originally adapted for. Physical systems are composed of multi-property components having several properties in different domains (mechanical, chemical, electrical, etc.) \cite{Taylor.2019}. E.g., a multifunctional enzyme has multiple properties in the same domain, which can produce expansive novelties whereas transformational novelties can be achieved by properties in different domains \cite{Taylor.2019}. The second is via \textit{non-additive composition}, which is phenotype generation by assembling several components drawn from a set of component types \cite{Taylor.2019}. E.g., the construction of proteins from amino acid sequences, producing new molecules, introducing new functions \cite{Taylor.2019}.

This assembly of lower-level elements into higher-level structures is also highlighted by Banzhaf et al. \cite{Banzhaf.2016}. With the above-mentioned metamodel that defines novelties, they also provide an abstract way to model multiple levels accounting for such constructed structures at different levels \cite{Banzhaf.2016}. They also mention that having several levels drastically increases the combinatorial possibilities to construct new structures and with that also the demand for computational power \cite{Banzhaf.2016}. It therefore seems to be a way to increase the opportunities to create something new. It also implies that open-ended evolution in computational systems is computational expensive.

\subsection{Modelling Contributions from Complex Systems Science}
There are also relevant modelling contributions from the field of complex systems science. W. Brian Arthur is known for his work on complexity economics \cite{Arthur.2014} and technology evolution \cite{ArthurSydney.2018,Arthur.2009}. He proposed the concept of \textit{combinatorial evolution}, which states that new technologies are created out of existing technologies and iteratively, these newly created technologies become building blocks for yet further technologies \cite{Arthur.2009,Arthur.2009qgi}. Technology is therefore self-creating or autopoietic \cite{ArthurSydney.2018,Arthur.2009qgi}. In a simple computer model of circuits, Arthur and Polak \cite{Arthur.2006} showed that complicated technologies (in their case circuits) could be created out of simpler building blocks, and they found evidence of self-organised criticality. It requires some kind of modularity and the evolution of simpler steppingstone technologies \cite{ArthurSydney.2018,Arthur.2009,Arthur.2009qgi,Arthur.2006}. The latter means that we cannot create a technology ahead of time without first creating the simpler precursor technologies. Natural phenomena also provide technological elements which can be combined \cite{ArthurSydney.2018,Arthur.2009qgi}. In terms of open-endedness, there seems to be a vast space of possible combinations and with the conversion of discovered natural phenomena into technological elements, there is a mechanism in place to expand that space.

Combinatorial evolution is also part of a more general approach to modelling evolution by the complex system scientist Stefan Thurner. He and his colleagues recently introduced the co-evolutionary, combinatorial, and critical evolution model (CCC model) \cite{ThurnerSydney.2018,Thurner.2018,Thurner.2010,Klimek.2012,Klimek.2010,Hanel.2005}. It models evolution as an open-ended process of creation and destruction of new entities emerging from the interactions of existing entities with each other and with their environment \cite{Thurner.2018}. The spaces of entities and of interactions co-evolve and new entities emerge spontaneously or through the combination of existing entities. This leads to power law statistics in histories of events \cite{Thurner.2018,Thurner.2010}. Selection is modelled by specifying rules for what can be created and what will be destroyed \cite{ThurnerSydney.2018,Thurner.2018,Thurner.2010}. This model captures so-called \textit{punctuated equilibria} in biological evolution \cite{Gould.1977} or \textit{Schumpeterian business cycles} in economic evolution \cite{Schumpeter.1939}, where an equilibrium is destabilised or destroyed by a critical transition leading to another equilibrium in an ongoing and thus open-ended process \cite{ThurnerSydney.2018,Thurner.2018,Thurner.2010,Thurner.2011}. It is interesting to note that it could be shown that in economic innovation, \textit{creative deconstruction} is happening and not \textit{niche filling} as usually assumed in biological innovation \cite{ThurnerSydney.2018,Thurner.2018,Klimek.2012}.

\subsection{Modelling Contributions from Artificial Intelligence and Evolutionary Algorithms}
Open-ended evolution is also studied in artificial intelligence and evolutionary algorithms. It is an emerging topic where the research of Kenneth O. Stanley serves as an example here. He tried to get rid of the prevailing concept to reward optimising a fitness function and has even suggested to abandon objectives in general \cite{Stanley.2015,Lehman.2011,Stanley.2010}. He showed that a novelty-driven approach finds solutions faster and results in solutions with less genomic complexity in comparison to traditional evolutionary computation \cite{Woolley.2014}. He also devised several algorithms including novelty search with explicit novelty pressure, MAP-Elites and innovation engines with explicit elitism within niches in an otherwise divergent process, and minimal criterion co-evolution where problems and solutions can co-evolve divergently \cite{Auger.2019,Bosman.2017}. Like Thurner, avoiding objectives also allowed Stanley to model punctuated equilibria with transitions between equilibria in a simple simulation with voxel structures \cite{Pugh.2017}. Also, in this case co-evolution and the apparently never-ending creation of anything new by combining existing structures were essential ingredients.

\subsection{Modelling Contributions from Evolutionary Biology}
The work of Thurner and Stanley indicates that transitions between equilibria are an important part of open-ended evolution. In evolutionary biology, the major evolutionary transitions are of great interest too, for example the transition from unicellular to multicellular organisms \cite{Szathmary.2015,Szathmary.1995}. Here, only a small selection of research is presented, mainly on mechanisms which can explain rapid increases in diversity and biological innovation. The work of evolutionary ecologist Ole Seehausen illustrates this well as he is interested in mechanisms by which diversity arises. Especially relevant here is the possibility of speciation through combinatorial mechanisms. In such cases, new combinations of old gene variants can quickly generate reproductively isolated species and thus provide a possible explanation for rapid speciation \cite{Marques.2019}. E.g., he showed that hybridisation between two divergent lineages provides ample genetic starting variation. This is then combined and sorted into many new species fuelling rapid cichlid fish adaptive radiations \cite{Meier.2017}. Seehausen furthermore investigates and underlines the importance of jointly considering species traits and environmental factors in speciation and adaptive radiation as they affect one another \cite{Seehausen.2009,Wagner.2012}. His work therefore supports the importance of co-evolutionary and combinatorial dynamics for open-ended evolution, even though co-evolution is between species in a heterogeneous environment and combinations happen at the gene level.

Biological insights into innovation itself are also relevant. The work of evolutionary biologist Andreas Wagner illustrates this nicely \cite{Hochberg.2017,Wagner.2011}. E.g., he showed that recombination creates phenotypic innovation in metabolic networks much more readily than random changes in chemical reactions \cite{Hosseini.2016}. The work of Wagner suggests that recombination of genetic material is a general mechanism which greatly increases the diversity of genotypes \cite{Wagner.2011je,Martin.2009}. Also, relevant here is his work on evolutionary innovation through exaptation. He found that simulated real metabolic networks were not only able to metabolise on a specific carbon source but also on several others, which shows that metabolic systems may harbour hidden pre-adaptations that could potentially lead to evolutionary innovations \cite{Barve.2013}. Combinatorial interactions at the gene-level again play a crucial role and the latter study revealing hidden pre-adaptations is like Stanley's open-ended algorithms creating many potential solutions before applied to solve an actual problem.

Besides this limited and biased review of contributions from evolutionary biology, it seems nevertheless important to point out that the field has shown that combinatorial interactions matter at organisational levels above the genes and that a changing environment can greatly affect species diversity and vice versa.

\section{Recent Advances in Implementing Open-Ended Evolutionary Systems}

\subsection{Implementation Contributions from Artificial Life and Open-Ended Evolution Community}

We first consider implementations from the artificial life and open-ended evolution community. Banzhaf et al. \cite{Banzhaf.2016} and Taylor \cite{Taylor.2019} provide some implementation suggestions. The implementation of computational systems which can detect and integrate novelties into the model and metamodel as described by Banzhaf et al. \cite{Banzhaf.2016} and Taylor \cite{Taylor.2019} provides a challenge in its own right. It is argued that operations should be defined \textit{intrinsically} in the system and by the system itself \cite{Taylor.2019,Packard.1988}. It requires program code which can recognise and modify itself. Banzhaf et al. \cite{Banzhaf.2016} state that this can be achieved by representing entities as strings of assembly language code, or by using a high-level language specifically designed for this purpose \cite{Spector.2002}, or a \textit{reflective} language. A reflective language allows implementing programs that have the ability to manipulate and observe their states during their own execution \cite{Demers.1995}. Indeed, it has been possible to generate exploratory, expansive, and transformative novelty with Stringmol, where modifications happen in sequences of assembly language code \cite{Stepney2020:ALife:innov}. A replicator and some of the observed operations and structures were extrinsically defined whereas some others could be defined intrinsically \cite{Stepney2020:ALife:innov}.

There are several computational systems of which Avida \cite{Ofria.2004} and Geb \cite{Channon.2010} are two prominent examples. Usually digitally simulated organisms are represented by assembly code competing for limited CPU resources. Most of these systems extrinsically implement common shortcuts such as replication and a certain fitness function, which makes them a powerful tool to explore biological questions such as the genotype-phenotype mapping \cite{Fortuna.2017z77} in a highly controlled way. Another strength of computational systems is that they usually involve visualisations, e.g., Sims \cite{Sims.1994}, and with that help exploring complex evolutionary dynamics. With respect to open-endedness, however, Pugh et al. \cite{Pugh.2017} point out that none of these systems has generated explosions of complexity, as seen in biological evolution during transitions and therefore something must still be missing. With Voxelbuild, Pugh et al. \cite{Pugh.2017} contributed the most relevant computational system in this respect. A first prototype demonstrated that a certain organisation of voxels emerged which was used as a steppingstone for yet other organisations appearing later. This seems to be like combinatorial evolution. Additionally, they report that exaptation occurred, which reminds of the evolutionary biology studies.

\subsection{Implementation Contributions from Complex Systems Science}

Thurner et al. \cite{Thurner.2018} add another important aspect to the implementation. They argue that only a so-called \textit{algorithmic} implementation and thus discrete formulation can work because in evolutionary systems, boundary conditions cannot be fixed (the environment evolves as a consequence of the system dynamics), and the phase-space is not well defined as it changes over time \cite{Thurner.2018}. It would lead to a system of dynamical equations that are coupled dynamically to their boundary conditions, which is according to them a mathematical monster and the reason why evolutionary systems cannot be implemented following an analytical approach. In addition, with the CCC model, they provide a general description of a complex evolving system that is so general that it applies to every evolutionary system. It therefore provides a general metamodel layer of a computational evolutionary system, which suggests that it might not need a change to capture novelties.

\section{The Allagmatic Method}

\subsection{Modelling Contribution}

We have developed the so-called \textit{allagmatic method} \cite{Christen.2019,Christen.2021b} to describe, model, implement, and allow interpretation of complex systems. It consists of a system metamodel inspired and guided by philosophical concepts of Gilbert Simondon \cite{Simondon.2020,Simondon.2017} and Alfred North Whitehead \cite{Debaise.2017,Whitehead.1978}. Simondon's metaphysics gives an operational and systemic account of how technical and natural objects emerge and evolve. It allows abstractly defining a system with the concepts \textit{structure} and \textit{operation} since according to him, systems develop starting with a seed by a constant interplay between operations and structures \cite{DelFabbroDiss.2021}. More concretely, but still general, we defined model building blocks in a system metamodel. The main building blocks are \textit{entity}, \textit{milieu}, \textit{update function}, \textit{adaptation function}, and \textit{target}, for which we recently provided a mathematical formalism \cite{Christen.2020o7}. The concepts entity, adaptation, and control are borrowed from Whitehead \cite{Debaise.2017,Whitehead.1978} as described in a recent technical report \cite{DelFabbro.2020}. The creation of a system model and metamodel can be followed through three regimes: In the virtual regime, abstract definitions with classes corresponding to interpretable philosophical concepts are given. Using generic programming \cite{Czarnecki.2000}, the type of states an entity can have, are defined by defining a system model object with no parameters initialised yet. At this point the metastable regime starts, where step by step the object/model is concretised with parameters such as number of entities and concrete update functions (model individuation). Once all parameters are defined, the object is executed in the actual regime. If there are any adaptation processes involved, the allagmatic method cycles between the metastable and actual regimes.

\subsection{Implementation Contribution}

The programming of the allagmatic method with its system metamodel is aligned with philosophical concepts. This not only allows interpretation of the final result in the context of the related metaphysics, but it also allows to follow the developmental steps a model is undergoing and thus provides a way to study the emergence of typical characteristics of complex systems. We recently outlined how adaptation can be studied in this way in a working paper \cite{DelFabbro.2020}. There, we also introduced the possibility and concepts to form hierarchies and define control, which further supports the use of the allagmatic method to define concepts that are difficult to pin down.

Furthermore, we showed how the method might be used for automatic programming \cite{Christen.2021b}. We found that the abstract model building blocks are well suited to be automatically combined by self-modifying code in a high-level language. Our work shows that certain philosophical concepts and even metaphysics as a whole can be defined and implemented in program code providing the opportunity to run these concepts or the whole metaphysics and study them in action.

We also created a prototype of open-ended automatic programming through combinatorial evolution \cite{Fix.2021}. Like Arthur and Polak \cite{Arthur.2006}, we created a computational model based on combinatorial evolution but instead of evolving circuits, we evolved computer code. Useful code blocks were stored in a repository and could be used in later iterations. Starting with basic keywords available in the programming language, more complex code blocks including classes, void methods, and variable declarations evolved.

\section{Current Modelling and Implementation Challenges}

\subsection{Modelling Open-Ended Evolutionary Systems}

\textit{Co-evolutionary dynamics}, \textit{combinatorial interactions}, and a \textit{changing environment} seem to be important ingredients of open-ended evolutionary systems. The work of evolutionary biologists including Seehausen and Wagner supports the view that co-evolutionary dynamics and combinatorial interactions are key elements. They also indicate that biological evolution exhibits different levels or types of combinatorial interactions, and that the environment is an important driver and mediator of change. The CCC model accounts for co-evolutionary dynamics and combinatorial interactions, and successfully generates the statistics of economic data with ever reoccurring transitions between equilibria \cite{ThurnerSydney.2018,Thurner.2018,Thurner.2010,Klimek.2012,Klimek.2010,Hanel.2005}. It could also show that economic innovations are driven by creative destruction, thus Schumpeterian evolution. This provides important insights into open-ended dynamics of economic evolution  \cite{ThurnerSydney.2018,Thurner.2018,Klimek.2012}. However, it still needs be investigated in other evolutionary systems, especially in biological evolution. Banzhaf et al. \cite{Banzhaf.2016} and Taylor \cite{Taylor.2019} provide guidance for modelling open-ended evolution in general which might allow us to come up with a model that captures open-ended dynamics of any evolutionary system, including economic and biological systems.

\subsection{Implementing Open-Ended Evolutionary Systems}

There is the challenge of an intrinsic implementation of open-ended evolutionary systems. The programming techniques already exist to do that; however, the real challenge is linking the structure and events in the implementation with interpretable concepts. To illustrate this problem, let us assume giving up shortcuts completely and letting the program overwrite the model and metamodel completely. Having no replicator or other prevailing concepts makes it hard to understand and see what is going on in the evolutionary simulation. This problem was discussed at the third workshop on open-ended evolution \cite{Packard.201955,PackardI.2019}. It is mostly uncharted territory needing much more research, including how to identify certain concepts and components from simulation data and how to implement such systems in a purely intrinsic manner, where generated novelties are meaningfully integrated into the model/metamodel by the evolving systems themselves.

Another challenge is the choice of digital organisms and environment. The CCC model \cite{ThurnerSydney.2018,Thurner.2018,Thurner.2010,Klimek.2012,Klimek.2010,Hanel.2005} provides a mathematical formalism for theoretical considerations and ways to perform statistical analyses. Computational systems from artificial life and open-ended evolution community such as Voxelbuild \cite{Pugh.2017} usually come with powerful visualisations, however, they lack a mathematical formalism.

\section{The Allagmatic Method for Open-Ended Evolutionary Systems}

\subsection{Model Building Blocks of Open-Ended Evolutionary Systems}

Observing evolving systems like technology or the rain forest makes clear that not only entities evolve but also interactions among them. Co-evolution implies that species affect each other reciprocally \cite{Debaise.2012}. Since species are also part of the environment, co-evolution leads to a changing environment providing more possibilities for state changes. Also, external environmental input can change and effect species and their interactions. Combinatorial interactions create new entities from existing entities \cite{ArthurSydney.2018,Arthur.2009,ThurnerSydney.2018,Thurner.2018}. These newly created entities might be able to exploit different parts of the changing environment and with that might be able to fill niches arising. Chromaria \cite{Soros.2018} captures this to some degree as entities become part of the environment and thus change the environment that interacts with further new entities. Changing the interactions between entities and between entities and their environment leads to complex cascades of changes potentially leading to disruptive changes in the system that can be regarded as novelties. Combinatorial interactions also lead to evolutionary changes and potential novelties, they combine existing entities to form new entities. This can be nicely observed in the evolution of technology \cite{ArthurSydney.2018,Arthur.2009}. It is a possibility how transitions might be explained, e.g., from unicellular to multicellular organisms \cite{Szathmary.2015,Szathmary.1995}.

One could therefore use the allagmatic method to capture co-evolutionary dynamics including with the environment and combinatorial interactions as given by the CCC model. The CCC model has been able to generate an ongoing evolutionary process with punctuated equilibria when in addition the lifetime of an entity was limited \cite{ThurnerSydney.2018,Thurner.2018,Thurner.2010}. It showed the statistical behaviour of open-ended evolutionary systems. Here, the CCC model is formalised within the allagmatic method to allow interpretation within the implemented metaphysics of Simondon \cite{Simondon.2020,Simondon.2017} and Whitehead \cite{Debaise.2017,Whitehead.1978}. The system metamodel of the allagmatic method and the CCC model both follow a complex systems perspective, which makes them compatible.

The model building blocks or general concepts to be captured with the allagmatic method are specifically: evolving entities, entity lifetime parameters, co-evolutionary operations of entities and environment, and combinatorial interactions.

\subsection{System Model and Metamodel of Open-Ended Evolution}

The allagmatic method consists of a system metamodel for modelling systems in general and complex systems in particular (see Christen \cite{Christen.2020o7} for detailed mathematical definitions). The system metamodel describes individual parts of a system as entities defined with an entity $e$-tuple $\mathcal{E}=(\hat{e}_1,\hat{e}_2,\hat{e}_3,\dots,\hat{e}_{e})$, where $\hat{e}_i\in Q$ with $Q$ being the set of $k$ possible entity states. Entity states are updated over time according to an update function $\phi : Q^{m+1} \rightarrow Q$ with $m$ being the number of neighbouring or linked entities. The update function $\phi$ therefore describes how entities evolve over time dependent on the states of neighbouring entities. Update rules and thus the logic are stored in the structure update rules $\mathcal{U}$. Entities are thereby considered connected together in a network structure and defined with the milieus $e$-tuple $\mathcal{M}=(\hat{\mathcal{M}}_1,\hat{\mathcal{M}}_2,\hat{\mathcal{M}}_3,\dots,\hat{\mathcal{M}}_{e})$, where $\hat{\mathcal{M}}_{i}=(\hat{m}_1,\hat{m}_2,\hat{m}_3,\dots,\hat{m}_m)$ is the milieu of the $i$-th entity $\hat{e}_i$ of $\mathcal{E}$ consisting of $m$ neighbours of $\hat{e}_i$. Over time, update function $\phi$ and milieus $\mathcal{M}$ might be changing as well, which is described with the adaptation function $\psi$. 

We now extend the system metamodel where needed with concepts to model open-ended evolution as identified above. \textit{Evolving entities:} The entity $e$-tuple $\mathcal{E}$ captures evolving entities (entities changing their state over time) in the same sense as the general evolution algorithm \cite{Thurner.2018} does with the state vector $\sigma$. The general evolution algorithm can be regarded as the metamodel from which the CCC model is created \cite{Thurner.2018}. \textit{Co-evolutionary operations of entities and environment:} With the creation of new entities (novelty), also new possibilities for interactions emerge. This is key for open-endedness and is captured by the co-evolution of entities and their interactions in the general evolution algorithm \cite{Thurner.2018}. Formally, the update equations of the entity state vector $\sigma$ and the interaction tensors $M$ are simultaneously updated over time in the general evolution algorithm. In the system metamodel of the allagmatic method, this is described with the update function $\phi$ and the adaptation function $\psi$ that can both be modelled in such a way that the update function $\phi$ updates entity states in $\mathcal{E}$ simultaneously with the adaptation of their interactions in the milieus $\mathcal{M}$ through the adaptation function $\psi$. This is a concretisation of the system metamodel into a metamodel of open-ended evolution. The environment is also part of co-evolution and described in the state vector $\sigma$ in the general evolution algorithm \cite{Thurner.2018} and the entity $e$-tuple $\mathcal{E}$ in the system metamodel. \textit{Combinatorial interactions:} In complex systems, interactions are of combinatorial nature consisting of rules determining how new entities can be formed out of existing entities. The creation and destruction of entities is encoded in rules that do not change with time. They can be regarded as physical or chemical laws determining which transformations and reactions are possible, respectively. Please note that this covers the typical evolutionary mechanisms of selection, competition, and reproduction. At runtime, models created from this metamodel make use only of a subset of these rules at any given time point, which is captured with so-called active productive/destructive rules. This is formally described with the function $F$ in the general evolution algorithm \cite{Thurner.2018} and the update function $\phi$ in the system metamodel. \textit{Entity lifetime parameter:} Besides the creation of new entities through combinatorial interactions, entities can spontaneously appear, which would be like discovering a new law or element in nature. By introducing a decay rate $\lambda$, the CCC model did not freeze \cite{Thurner.2018}. It thus plays an important role for open-ended evolution. In the system metamodel, this parameter can be described as a further structure with a respective further operation. 

\subsection{Self-Modifying Code Prototype in C\#}

An intrinsic implementation as suggested by Banzhaf et al. \cite{Banzhaf.2016} and Taylor \cite{Taylor.2019} requires self-modifying program code and some way to add novelties to the model or metamodel. Interpreting these novelties in the context of a certain metaphysical framework will most likely require a high-level language with the capabilities to modify program code during runtime and reflect on it. C\# provides these capabilities with the open-source Roslyn .NET compiler \cite{DotNETCompilerPlatformSDK}. The compiler platform provides dynamic code manipulation with syntax trees and many other features including reflection as well as comprehensive code analysis. Syntax trees can either be created from a string containing program code or they can be assembled using predefined classes. As opposed of writing program code into a file and then compile it, syntax trees can be stored as an object and compiled and executed at runtime.

In the allagmatic method, a general layer in the system metamodel that is not modifiable by the code is suggested here. These are the model building blocks every complex evolutionary system requires. However, there is also a layer in the metamodel that is modifiable by the code. It consists of less general model building blocks that are basically more concrete instances of the general layer. With these different layers and controllable code self-modification, it will potentially be possible to link concepts defined in the metamodel to newly generated code improving interpretability. The present study provides a first prototype of self-modifying code in C\# \cite{Skeet.2019} bringing us one little step closer to that ambitious goal.

Fundamentally, implementing self-modifying code requires considering at least three basic questions: what words to use, how to concatenate these words to create valid code, and how to implement the duality between code and data? In theoretical computer science, a \textit{word} or string is defined as a finite sequence of symbols over a given alphabet \cite{Kulkarni.2013}. An \textit{alphabet} is a finite set of symbols \cite{Kulkarni.2013} and \textit{symbols} are the basic constituents of any language (i.e., the set of all words over a given alphabet), e.g., letters, digits, or any other characters \cite{Kulkarni.2013}. To define the words, we consider universal computation and code interpretability. We want to choose words that do not limit the generated code and therefore require universal computation or Turing completeness. It has been shown that only the instructions load, store, increment, and goto (unconditional branching) are required to achieve universal computation \cite{Rojas.1996}. Most widely used programming languages including C++ and C\# provide words to generate these instructions and many other instructions and are thus capable of universal computation. Regarding code interpretability, we suggest including the complete or most of the syntax of a high-level programming language since these languages are designed to be human readable and interpretable. The first words to include are therefore all the C\# keywords as defined in the C\# language reference \cite{CSDocKeywords}. In addition to keywords, we also include special characters, the member access expression, and operators as defined in the C\# language specification \cite{CSLangSpec} as well as some further words. Please note that we treat symbols such as operators as words since one can concatenate them together with other words to create sentences (i.e., instructions). In the following, all included words are listed:

\begin{itemize}
  \item Keywords: \texttt{abstract}, \texttt{as}, \texttt{base}, \texttt{bool}, \texttt{break}, \texttt{byte}, \texttt{case}, \texttt{catch}, \texttt{char}, \texttt{checked}, \texttt{class}, \texttt{const}, \texttt{continue}, \texttt{decimal}, \texttt{default}, \texttt{delegate}, \texttt{do}, \texttt{double}, \texttt{else}, \texttt{enum}, \texttt{event}, \texttt{explicit}, \texttt{extern}, \texttt{false}, \texttt{finally}, \texttt{fixed}, \texttt{float}, \texttt{for}, \texttt{foreach}, \texttt{goto}, \texttt{if}, \texttt{implicit}, \texttt{in}, \texttt{int}, \texttt{interface}, \texttt{internal}, \texttt{is}, \texttt{lock}, \texttt{long}, \texttt{namespace}, \texttt{new}, \texttt{null}, \texttt{object}, \texttt{operator}, \texttt{out}, \texttt{override}, \texttt{params}, \texttt{private}, \texttt{protected}, \texttt{public}, \texttt{readonly}, \texttt{ref}, \texttt{return}, \texttt{sbyte}, \texttt{sealed}, \texttt{short}, \texttt{sizeof}, \texttt{stackalloc}, \texttt{static}, \texttt{string}, \texttt{struct}, \texttt{switch}, \texttt{this}, \texttt{throw}, \texttt{true}, \texttt{try}, \texttt{typeof}, \texttt{uint}, \texttt{ulong}, \texttt{unchecked}, \texttt{unsafe}, \texttt{ushort}, \texttt{using}, \texttt{virtual}, \texttt{void}, \texttt{volatile}, \texttt{while}
  \item Special characters: \texttt{\{}, \texttt{\}}, \texttt{(}, \texttt{)}, \texttt{[}, \texttt{]}, \texttt{"}, \texttt{;}, \texttt{,}
  \item Member access expression: \texttt{.}
  \item Arithmetic operators: \texttt{+}, \texttt{-}, \texttt{*}, \texttt{/}, \texttt{\%}
  \item Relational operators: \texttt{<}, \texttt{>}, \texttt{<=}, \texttt{>=}, \texttt{==}, \texttt{!=}
  \item Logical operators: \texttt{\&}, \texttt{\^}, \texttt{|}, \texttt{\&\&}, \texttt{||}
  \item Assignment operator: \texttt{=}
  \item Further words: \texttt{IDENTIFIER}, \texttt{NUMBER}, \texttt{PLACEHOLDER}
\end{itemize}

There are some further words that need to be explained. Program code contains words that are used as a name or identifier, e.g., for variables and classes. To account for hat, the word \texttt{IDENTIFIER} is included as a possible word. If the self-modifying code choses this word, an identifier is generated and inserted. Natural numbers are inserted in the same way replacing the word \texttt{NUMBER}. If the word \texttt{PLACEHOLDER} is chosen, an instruction that is a combination of words representing valid code is inserted. Such a placeholder allows generating a nested code structure, e.g., a variable declaration inside a method \cite{Fix.2021}. In this first prototype of self-modifying code, syntax to make use of the extensive .NET API is not included. Thus, the language used here is a subset of the C\# language. 

We now address the second question regarding the combination of the defined words to create valid code. The general algorithm to achieve this is based on the concept of combinatorial evolution as proposed by W. Brian Arthur \cite{Arthur.2009,Arthur.2009qgi} and already used in our earlier study to evolve programming concepts such as variable declarations and classes in Java \cite{Fix.2021}. The algorithm uses two data structures, a list \texttt{words} (set $W$) containing the above defined words and a list \texttt{codeBlocks} (set $C$) storing valid code blocks (sentences), which is initially empty ($C=\emptyset$). It is an iterative process in which several steps are repeated: 1) The first step is to generate a new code block. This is achieved by randomly selecting a given number of words from set $W$, which are then concatenated separated by a space. The number of words in a code block is set randomly between 2 and 8 as in the previous studies where combinatorial evolution was simulated \cite{Arthur.2006,Fix.2021}. If the chosen word is \texttt{PLACEHOLDER}, it is replaced by an already existing valid code block from set $C$ or by another word in case set $C$ is still empty. Similarly, if the chosen word is \texttt{IDENTIFIER}, it is replaced by a numbered identifier. If the chosen word is \texttt{NUMBER}, it is replaced by a randomly generated integer. 2) The second step is checking the validity of the newly generated code block. This is achieved by parsing the code block into a syntax tree, which is then analysed making use of the Roslyn API (the .NET Compiler Platform SDK) \cite{DotNETCompilerPlatformSDK}. A compilation object is created from the syntax tree \cite{CSharpSyntaxTreeClass} that is compiled at runtime avoiding time consuming read and write operations in file-based compilation. 3) In the third step, if the compilation is successful, the code block is added to set $C$. Running this algorithm to create code blocks for 1 million iterations revealed some valid code blocks including variable declarations \texttt{char identifier98242 ;}, scope definitions \texttt{\{ \}}, and combinations of the two \texttt{char identifier98242 ; \{ \}}.

The third question is how to implement a duality between code and data. We need to address this question because we want to execute the generated code as well as modify it. It also includes transferring the state of data such as the state of an entity from code to data and back to code again. Such transitioning between code and data allows implementing self-modifying operations on data structures, e.g., the update function $\phi$ and update rules $\mathcal{U}$, where the current state of an entity is required to be transferred from code to data and after self-modification and running of the code, back again into code as the updated state of the entity. There are certain programming languages where program code is also represented as data and thus can be manipulated as data. It is a language property often referred to as homoiconic and a prominent example is Lisp. However, it can also be achieved with C\#. If, for example, we want to transfer the value of the variable \texttt{input} from code to data, we can use the \texttt{String.Replace} \cite{ReplaceMethod} and \texttt{String.ToString} \cite{ToStringMethod} methods as follows: \texttt{code.Replace("input", input.ToString())}. Here, \texttt{code} is a string containing code as data, \texttt{"input"} represents the variable \texttt{input} in that code as data, and \texttt{input} represents the variable \texttt{input} in the code as code. The variable value of the latter is converted into a string with the \texttt{String.ToString} method and then this value is inserted into the code as data by replacing \texttt{"input"} with the \texttt{String.Replace} method. We can then run the code as data (the string \texttt{code}) as described above in the self-modifying code prototype. Once we have executed the code at runtime, we also want to transfer back the output to our program, therefore from data to code. One way to achieve that in C\# is to redirect the console output stream to a variable. We first set the output stream to a \texttt{StringWriter} object \cite{StringWriterClass}. When the code as data (the string \texttt{code}) is executed, the value of the variable \texttt{output} is printed out in the console with the \texttt{Console.WriteLine} method \cite{WriteLineMethod} as follows: \texttt{Console.WriteLine(output)}. Because of the redirection of the output stream, \texttt{output} is not printed in the console but stored in the \texttt{StringWriter} object. From there we can use it in our program and thus we have transferred data to code.

%
%
%

\subsection{The Allagmatic Method as Guidance to Create Code Blocks}

The system metamodel $\mathcal{SM}$ of the allagmatic method is defined as:

\begin{equation}
		\mathcal{SM} \coloneqq (\mathcal{E},Q,\mathcal{M},\mathcal{U},\mathcal{A},\mathcal{P},\dots,\hat{s}_s,\phi,\psi,\dots,\hat{o}_o),
	\end{equation}
	
where $\mathcal{E}$ is the entities $e$-tuple, $Q$ the set of possible entity states, $\mathcal{M}$ the milieus $e$-tuple, $\mathcal{U}$ the update rules $u$-tuple, $\mathcal{A}$ the adaptation rules $a$-tuple, $\mathcal{P}$ the adaptation end $p$-tuple, $\hat{s}_s$ are further structures, $\phi$ the update function, $\psi$ the adaptation function, $\hat{o}_o$ are further operations, and $\hat{s}_i\in S \land \hat{o}_j\in O$ \cite{Christen.2020o7}. It intertwines structures and operations to create (complex) systems at the most abstract level in the virtual regime. Every such system contains at least the structures entities $\mathcal{E}$, entity states $Q$, milieus $\mathcal{M}$, update rules $\mathcal{U}$, adaptation rules $\mathcal{A}$, adaptation end $\mathcal{P}$, and the operations update function $\phi$, and adaptation function $\psi$. As stated above, to create an open-ended evolutionary system, we need the further structure lifetime parameter $\hat{s}_{lt}$ and a respective further operation $\hat{o}_{lt}$ to remove entities as soon as they have reached that time limit. We also need a simultaneous update of entity states and $\mathcal{U}$ and $\mathcal{M}$. Since the update function $\phi$ modifies entity states and the adaptation function $\psi$ modifies $\mathcal{U}$ and $\mathcal{M}$, this is achieved by running $\phi$ and $\psi$ in the same iteration. From the virtual regime, structures and operations are concretised creating a metastable system in the metastable regime. In this concretisation process, the self-modifying code prototype can now be used to generate concrete instances of the identified structures and operations. In the following, a description of how this could be implemented is provided for each structure and operation:

\begin{itemize}
  \item Entities $\mathcal{E}$ and their possible states $Q$: entities are concretised by defining the number of entities $e$, and number and kind of possible states $Q$ and $k$. By providing respective words, the self-modifying code is guided or restricted in its generation of code blocks. An entity is implemented as an object containing fields capturing all its states. E.g., it is possible to randomly define the number of entities and states and based on that, randomly chose a data type for each field from predefined words implementing different data types. A list of length $e$ is then created with the defined entity objects.
  \item Milieus $\mathcal{M}$: milieus are concretised by defining the number of connected entities $m$ for each entity and the connections between entities forming a network structure. E.g., it is possible to randomly define the number of connected entities for each entity and which entities are connected to it.
  \item Update function $\phi$ and update rules $\mathcal{U}$: update function and rules are concretised by defining what operations should be performed on which entity fields under which conditions. E.g., it is possible to randomly define some boolean operations for boolean fields and arithmetic operations for integer and float fields. These operations are then coupled with some randomly defined if-conditions choosing states of connected entities and relational operators.
  \item Adaptation function $\psi$, adaptation rules $\mathcal{A}$, and adaptation end $\mathcal{P}$: adaptation function, rules, and end are concretised by defining new update rules $\mathcal{U}$, milieus $\mathcal{M}$, and adaptation end $\mathcal{P}$. E.g., it is possible to overwrite update rules with newly created rules as described above, overwrite the network structure as described above, and randomly chose a value for each entity field to define the adaptation end.
  \item Lifetime parameter $\hat{s}_{lt}$ and respective operation $\hat{o}_{lt}$: lifetime parameter and operation are concretised by defining a time limit in terms of iterations and the way entities are removed. E.g., it is possible to randomly define an integer value for the lifetime parameter and to decide whether entities are removed after they were present for the defined number of iterations or by removing them randomly after the defined number of iterations passed.
\end{itemize}

Also, modification of code can be implemented with this approach by allowing overwriting certain parts of these definitions.

\section{Discussion and Conclusion}

Based on recent advances, the model building blocks \textit{evolving entities}, \textit{entity lifetime parameter}, \textit{co-evolutionary operations of entities and environment}, and \textit{combinatorial interactions} are identified to characterise open-ended evolutionary systems. These concepts led to punctuated equilibria in the CCC model \cite{Thurner.2018}, which means that it never reaches an equilibrium state and thus can be regarded open-ended. This study provides a formal description of a system metamodel for open-ended evolution according to the CCC model thus also capable of generating punctuated equilibria. It also provides a self-modifying code prototype in C\# and guidance to create code blocks that potentially will allow an intrinsic implementation of open-ended evolutionary systems as suggested by Banzhaf et al. \cite{Banzhaf.2016} and Taylor \cite{Taylor.2019}.

The proposed self-modifying code prototype and the guidance of the allagmatic method to create blocks seem to be a promising way to change code at runtime and potentially account for novelties. This is achieved by controlling the self-modification of code within abstractly defined building blocks of a system metamodel describing complex and evolutionary systems in general. It could thus be a way to interpret novelties without limiting possible solutions.

It is interesting to note that certain models anticipate changes that might occur to them. In all evolutionary systems, new entities arise, and other entities disappear, which will not only change how many entities there are but also their interactions with each other and the environment. The CCC model \cite{Thurner.2018} is capable of accounting for such changes in the model through co-evolution of entity states and interactions. On this level, it therefore does not need self-modification of the code but generic programming \cite{Czarnecki.2000} of certain structures to dynamically adapt them to these changes.

In conclusion, interpretation of concepts within a metaphysical framework as described with the allagmatic method provides a promising starting point to interpret novelty generated at runtime. This study provides a system metamodel of open-ended evolution and a prototype of self-modifying code implemented in C\#. Using this prototype in the allagmatic method allows modifying certain structures and operations of the system model and metamodel in a controlled way and potentially will allow interpreting novelties with the help of the metaphysical framework.

%
%
\section*{Acknowledgement}
This work was supported by the Hasler Foundation under grant No.\ 21017. I thank Tom Van Dooren, Frietson Galis, Olivier Del Fabbro, Stefan Thurner, and Sagi Nedunkanal for helpful comments on a draft of the manuscript.
\bibliographystyle{unsrt}
\bibliography{References.bib}
\end{document}